\documentclass[letterpaper, 10 pt, conference]{ieeeconf}  

\IEEEoverridecommandlockouts                              

\usepackage{amsmath,bm,amsfonts}
\usepackage{graphicx}
\usepackage[font=small]{caption}
\usepackage[rightcaption]{sidecap} \sidecaptionvpos{figure}{c}
\usepackage{subcaption}
\usepackage{tabularx,booktabs}
\usepackage{dblfloatfix} 

\usepackage{algorithmicx}
\usepackage[ruled,vlined]{algorithm2e}
\usepackage[noend]{algpseudocode}
\algdef{SE}[DOWHILE]{Do}{doWhile}{\algorithmicdo}[1]{\algorithmicwhile\ #1}%

\title{\LARGE \bf
A Brain Inspired Learning Algorithm for the Perception of a Quadrotor in Wind
}

\author{Ajith Anil Meera$^{1}$ and Martijn Wisse$^{2}$
\thanks{$^{1,2}$ are with Cognitive Robotics, Faculty of Mechanical Maritime and Materials Engineering, Delft Institute of Technology, The Netherlands, Corresponding author:
        {\tt\small ajitham1994@gmail.com}}%
}

\begin{document}

\maketitle
\thispagestyle{empty}
\pagestyle{empty}

\begin{abstract}

The quest for a brain-inspired learning algorithm for robots has culminated in the free energy principle from neuroscience that models the brain's perception and action as an optimization over its free energy objectives. Based on this idea, we propose an estimation algorithm for accurate output prediction of a quadrotor flying under unmodelled wind conditions. The key idea behind this work is the handling of unmodelled wind dynamics and the model's non-linearity errors as coloured noise in the system, and leveraging it for accurate output predictions. This paper provides the first experimental validation for the usefulness of generalized coordinates for robot perception using Dynamic Expectation Maximization (DEM). Through real flight experiments, we show that the estimator outperforms classical estimators with the least error in output predictions. Based on the experimental results, we extend the DEM algorithm for model order selection for complete black box identification. With this paper, we provide the first experimental validation of DEM applied to robot learning.

\end{abstract}

\section{INTRODUCTION}

The inferior uncertainty handling capabilities of modern robot algorithms when compared to a human brain has been inspiring roboticists to search for brain inspired algorithms for robot perception. The recent advancements in computational neuroscience has culminated in the Free Energy Principle (FEP) that models the brain's perception and action under one optimization scheme. Fundamentally rooted in Bayesian Inference, FEP emerges as a brain theory that can learn hierarchical causal dynamic models from limited data under uncertainties. In light of these developments, we aim to bridge the gap between FEP and robotics by providing the first experimental proof of concept for one of its variant called Dynamic Expectation Maximization (DEM), applied to robot model learning.

\begin{figure}[!t]
\centering
\includegraphics[scale = 0.3]{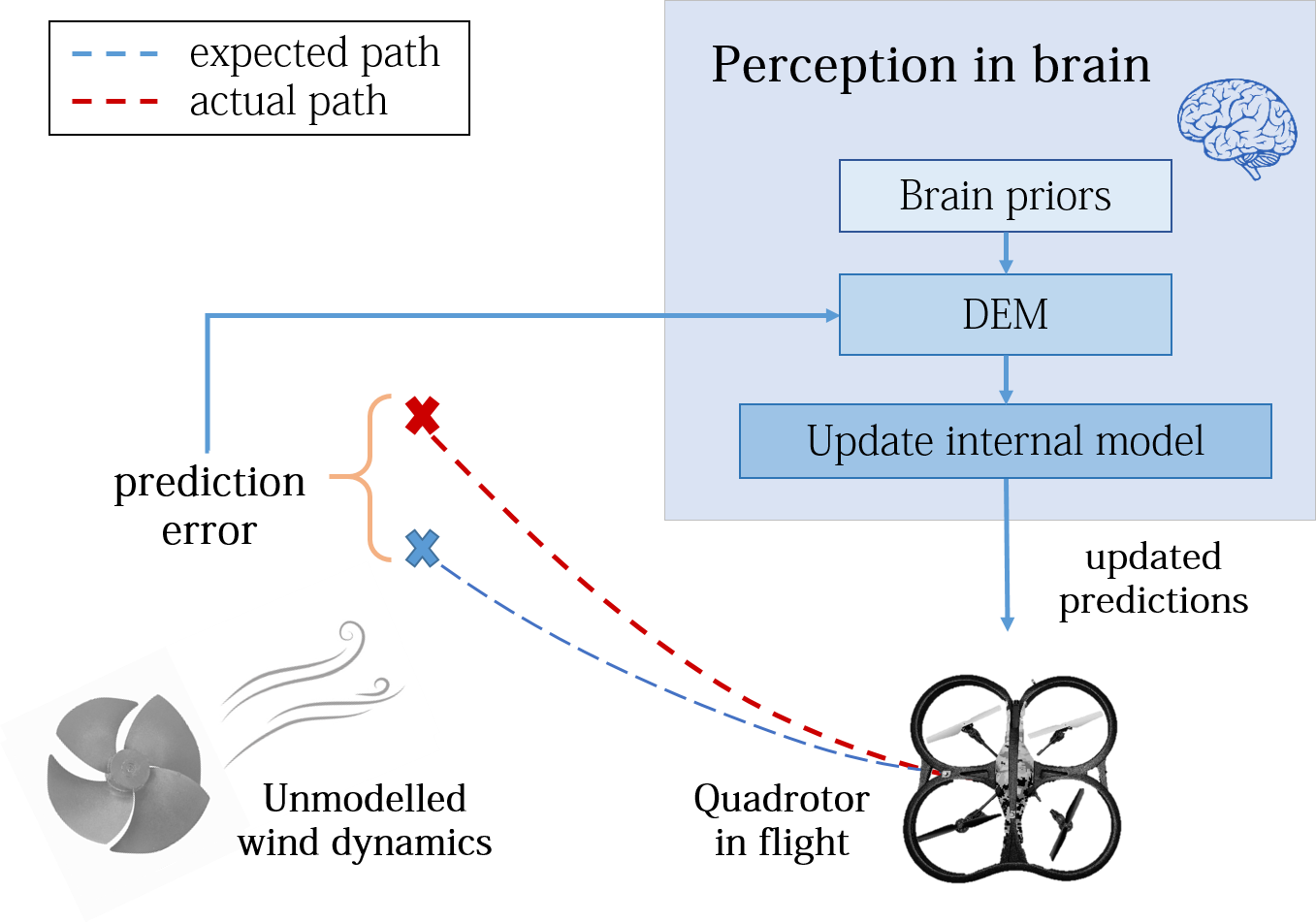}
\caption{The proposed DEM based perception scheme. The unmodelled wind and the non linearity errors in the robot brain's internal model manifests as the sensory surprisal (output prediction errors). The free energy principle (DEM) drives the robot brain to update its internal model of the world (quadrotor dynamics) for uncertainty resolution, resulting in better output predictions. }
\label{fig:drone_brain_diagram}
\end{figure}

With this work, we introduce the idea of leveraging the information content from the unmodelled system dynamics for accurate output predictions under uncertainty. We propose a DEM based perception scheme that models this noise (prediction error) as colored using generalized coordinates for accurate output predictions. Fig. \ref{fig:drone_brain_diagram} shows our proposed perception scheme applied to a quadrotor hovering under unmodelled wind conditions. The possible applications of this approach is wide and include the handling of unmodelled wind dynamics on a delivery drone, non-linearity errors of an industrial manipulator robot, friction dynamics of a skid steer ground robot in unknown terrains like martian surface, search and rescue environments etc. The core contributions of the paper include:
\begin{enumerate}
    \item introduce a DEM based perception scheme for the output prediction of a quadrotor hovering in wind,
    \item provide the first experimental confirmation for the usefulness of generalized coordinates for robot perception,
    \item extend the DEM algorithm with model order estimation for the black-box identification of linear systems.
\end{enumerate}

\section{RELATED WORK}
In this section, we capture the multidisciplinary nature of FEP literature, connecting neuroscience and robotics.
\subsection{Cognitive neuroscience}
FEP emerges from neuroscience as a unifying brain theory \cite{friston2010free} that explains the brain functions under a single framework - free energy optimization. According to FEP, every self organising system that is in equilibrium with the environment should minimize its free energy. This drives a biological agent into minimizing its sensory surprisal for uncertainty resolution while interacting with the environment. This is done in two ways - through perception (learning) and action (active inference). Perception involves learning the generative process in the environment for accurate predictions, whereas active inference involves acting on the environment to suppress sensory surprisal. Numerous methods have been developed around these ideas to explain brain functions including predictive coding \cite{friston2009predictive}, hierarchical brain models \cite{friston2008hierarchical}, active inference \cite{friston2011action}, DEM \cite{friston2008variational} etc. The biological plausiblity of FEP rests in its capability to provide a mathematical description of brain functions \cite{friston2009free}, to unify action and perception \cite{friston2010action}, to connect physiological constructs like memory, attention, value, reinforcement and salience \cite{friston2009free}, to explain active vision \cite{parr2021generative}, while remaining consistent with Freudian ideas\cite{carhart2010default}. Similarities of FEP with reinforcement learning \cite{friston2009reinforcement}, neural networks \cite{buckley2017free,friston2008hierarchical}, Kalman Filtering \cite{meera2020free}, PID control \cite{baltieri2019pid} and active learning \cite{friston2010action} further guides the quest for a brain inspired robot learning algorithm towards FEP as the unified robot learning algorithm. 

\subsection{Robotics}
Recent applications of FEP in robotics include the body perception of humanoid robots \cite{oliver2019active}, adaptive control for robot manipulators \cite{pezzato2020novel}, robot navigation and mapping (SLAM) \cite{ccatal2021robot} etc. Simultaneous state and input observer designs for linear time invariant (LTI) systems with colored noise was developed \cite{meera2020free} and applied to quadrotors \cite{ajith2021input_drone}. On the learning side, DEM was developed into a system identification method \cite{anilmeera2021DEM_LTI} with theoretical convergence guarantees \cite{ajith2021convergence}. In this paper we provide the proof of concept for DEM by learning a quadrotor model for output prediction.

\subsection{System identification}
In control systems, output predictions can be done using system identification, which is a mature field \cite{ljung1998system} with methods like Subspace (SS) method, Expectation Maximization (EM), Prediction Error Minimization (PEM). However, most of them consider the noises to be white, which is often a wrong assumption in practice and results in biased estimation for least square based methods \cite{zhang2011unbiased} and inaccurate convergence for the iterative methods \cite{liu2010least}. Although various bias compensation methods have been proposed to solve this problem \cite{zheng1998least,zhang2011bias}, none of them perform simultaneous state, input, parameter and noise hyperparameter estimation, except for DEM. Therefore, DEM is of importance to the research community and requires experimental validation on real robots. With this paper, we aim to fill this research gap.

\section{PROBLEM STATEMENT}
Consider the linearized plant dynamics given in Equation \ref{eqn:general_LTI} where $\textbf{A}$, $\textbf{B}$ and $\textbf{C}$ are constant system matrices, with hidden state $\textbf{x}\in \mathbb{R}^n$, input $\textbf{v}\in \mathbb{R}^r$ and output $\textbf{y}\in \mathbb{R}^m$.
\begin{equation} 
\label{eqn:general_LTI}
    \begin{split}
       \Dot{\textbf{x}} = \textbf{Ax}+\textbf{Bv} + \textbf{w},
    \end{split}
    \quad \quad
    \begin{split}
        \textbf{y} = C\textbf{x} +\textbf{z}.
    \end{split}{}
\end{equation}{}Here $\textbf{w}\in \mathbb{R}^n$ and $\textbf{z}\in \mathbb{R}^m$ represent the process and measurement noise respectively.  In this paper, we consider a special case where the system is a hovering quadrotor. Variables of the plant are denoted in boldface, while its estimates are denoted in non-boldface. The noises are assumed to be colored such that it was generated by the convolution of white noise with a Gaussian filter. The unmodelled wind dynamics and the non-linearity errors enter the system through $w$, making it colored. The problems considered in the paper are: 1) learn an LTI model to accurately predict the output $y$ from the inputs $v$, and 2) learn the order of the system $n$ for black box identification.

\section{PRELIMINARIES}
To lay the foundations of our perception and system identification scheme, this section introduces the key concepts behind DEM.

\subsection{Generalized coordinates}
The key concept that  differentiates DEM from other methods is its use of Generalized Coordinates (GC). GC is a relatively new concept in robotics and shouldn't be confused with the definition in multi-body dynamics. GC enables the estimator to gracefully handle colored noise by modelling the trajectory (instead of point estimates) of all the time dependent components like states, inputs, outputs and noises using their higher order derivatives, thereby providing additional information for perception. For example, the states in GC is given by $\tilde{x}=[x \ x' \ x'' \ ...]^T$. The variables in generalized coordinates are denoted by a tilde, and their components (higher derivatives) are denoted by primes. In Section \ref{sec:GC_imp} we will demonstrate the usefulness of GC in providing additional information for robot perception.

\subsection{Generative model}
The generative model denotes the robot brain's internal model of the generative process in the environment that is responsible for data generation. Since the time dependent components of the generative model are differentiable and because the noises are coloured, the evolution of states of an LTI system (generative process) in Equation \ref{eqn:general_LTI} can be extended as:
\begin{equation} 
    \begin{split}
        &x'=Ax+Bv+w \\
        &x''=Ax'+Bv'+w'\\ &...
    \end{split}
    \quad \quad
     \begin{split}
        &y=Cx+z \\
        &y' = Cx'+z'\\ &...
    \end{split}   
\end{equation}{} 
which can be compactly written as: \begin{equation}
\label{eqn:generative_process}
    \begin{split}
        &{\tilde{{x}}}'  = D^x\tilde{{x}}= \tilde{A}\tilde{{x}}+\tilde{B}\tilde{{v}} +\tilde{{w}} 
    \end{split}{}
    \quad \quad
    \begin{split}
        &\tilde{{y}} = \tilde{C}\tilde{{x}}+\tilde{{z}}
    \end{split}{}
\end{equation}{} where $D^x = \Bigg[ \begin{smallmatrix}{}
0 & 1 & & &\\
 & 0 & 1 & & \\
 & & .& . &  \\
 & & & 0& 1 \\
 & & & & 0
\end{smallmatrix}\Bigg]_{(p+1)\times (p+1)} \otimes I_{n\times n}$ \\ performs derivative operation, equivalent to shifting up all components in generalized coordinates by one block. $p$ and $d$ are the order of generalized motion of states and inputs respectively. Here, $\tilde{A}=I_{p+1}\otimes A$, $\tilde{B} = I_{p+1}\otimes B$ and $\tilde{C} = I_{p+1}\otimes C$, where $\otimes$ is the Kronecker tensor product. The generalized motion of output $\tilde{\textbf{y}}$ are computed from the discrete measurements $\textbf{y}$ \cite{meera2020free}.


\subsection{Colored noise modeling}
The colored noises are analytic such that the covariance of noise derivatives $\Tilde{z} = [z,{z'},{z''},...]^T$ and $\Tilde{w} = [w,{w'},{w''},...]^T$ are well defined. The correlation between noise derivatives are represented using the temporal precision matrix $S$ (inverse of covariance matrix). Since the correlation is assumed to be due to a Gaussian filter, $S$ becomes \cite{friston2008variational}:
\begin{equation}
\label{eqn:Smatrix}
    S(\sigma^2) = \begin{bmatrix}
    1 & 0 & -\frac{1}{2\sigma^2} & .. \\
    0 & \frac{1}{2\sigma^2} & 0 & ..\\
    -\frac{1}{2\sigma^2} & 0 & \frac{3}{4\sigma^4}& ..\\
    .. & .. & .. & ..
    \end{bmatrix}^{-1}_{(p+1)\times (p+1)}
\end{equation}
where $\scriptstyle \sigma^2$ is the variance of Gaussian filter, with $\sigma$ denoting the noise smoothness. While $\sigma^2=0$ denotes white noise, non-zero $\sigma^2$ denotes colored noise. The generalized noise precision matrices are given by $\tilde{\Pi}^w = S(\sigma^2)\otimes \Pi^w$ and $\tilde{\Pi}^z = S(\sigma^2)\otimes \Pi^z$, where $\Pi^w$ and $\Pi^z$ are the inverse noise covariances.

\subsection{Parameters and noise hyperparameters}
The parameter $\theta$ is composed of vectorized $A$, $B$, $C$ matrices defined as $\theta = \begin{bmatrix} vec(A^T)^T & vec(B^T)^T & vec(C^T)^T
\end{bmatrix}^T $, and the noise hyperparameters $\lambda = \begin{bmatrix} \lambda^z & \lambda^w \end{bmatrix}^T$ models the noise precision:
\begin{equation} \label{eqn:hyper_def}
    \Pi^w(\lambda^w) = e^{\lambda^w}\Omega^w, \ \Pi^z(\lambda^z) = e^{\lambda^z}\Omega^z,
\end{equation}
where $\Omega^w$ and $\Omega^z$ represent constant matrices encoding how different noises are correlated. We use $\Omega^w$ and $\Omega^z$ as identity matrices for this work. Parameter and hyperparameter estimation entails the estimation of $\theta$ and $\lambda$ respectively.

\subsection{Priors of the brain}
DEM enables the transfer of prior knowledge through Gaussian prior distributions for inputs, parameters and hyperparameters, centred around $\eta$ as $p(\tilde{v})= \mathcal{N}(\eta^{\tilde{v}},P^{\tilde{v}})$, $p(\theta) = \mathcal{N}(\eta^\theta,P^{\theta})$ and $p(\lambda) = \mathcal{N}(\eta^\lambda,P^{\lambda})$ respectively. The mean $\eta$ acts as the starting point for the perception on new data and the precision $P$ shapes the confidence on these priors. $P$ controls the robot brain's exploration-exploitation trade off during learning - lower $P$ favours exploration, whereas higher $P$ favours exploitation. We will exhaustively use this idea to pass known information to the algorithm (for example, known inputs ) through $\eta$ with high $P$.

\begin{figure*}[!htb]
    \centering
    \begin{subfigure}[b]{0.32\textwidth}  
		\includegraphics[width=\textwidth,trim={0.1cm 0 0.3cm 0},clip]{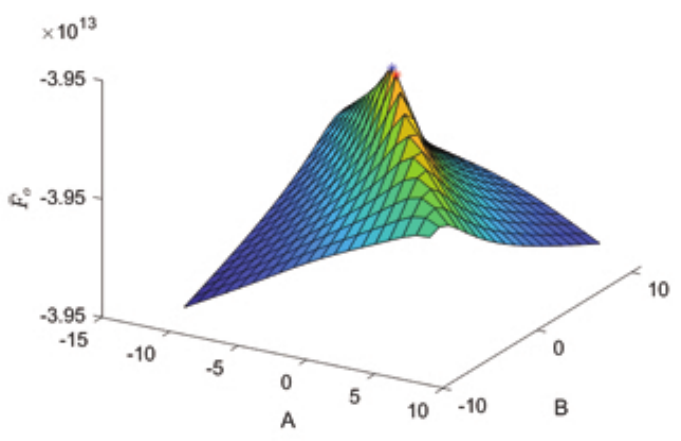}
		\caption{$\Bar{F}^o$ with respect to parameters $A$ and $B$.} \label{fig:f_bar}
    \end{subfigure} 
    \begin{subfigure}[b]{0.32\textwidth} 
		\includegraphics[width=\textwidth,trim={.1cm 0 .3cm 0},clip]{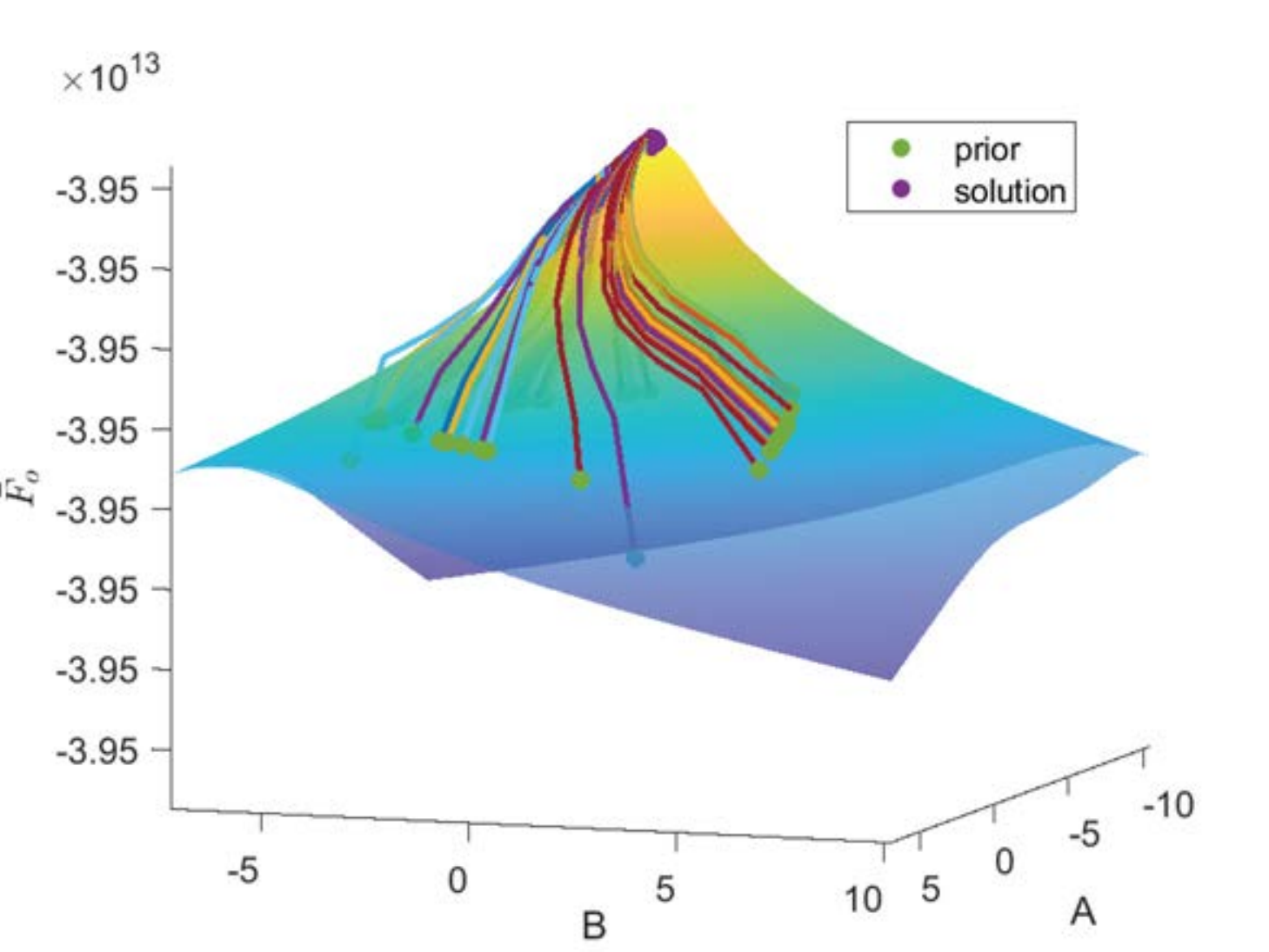}
		\caption{Perception as maximization of $\Bar{F}^o$. } \label{fig:climb_FE}
    \end{subfigure}    
    \begin{subfigure}[b]{0.32\textwidth} 	
		\includegraphics[width=\textwidth,trim={.1cm 0 .3cm 0},clip]{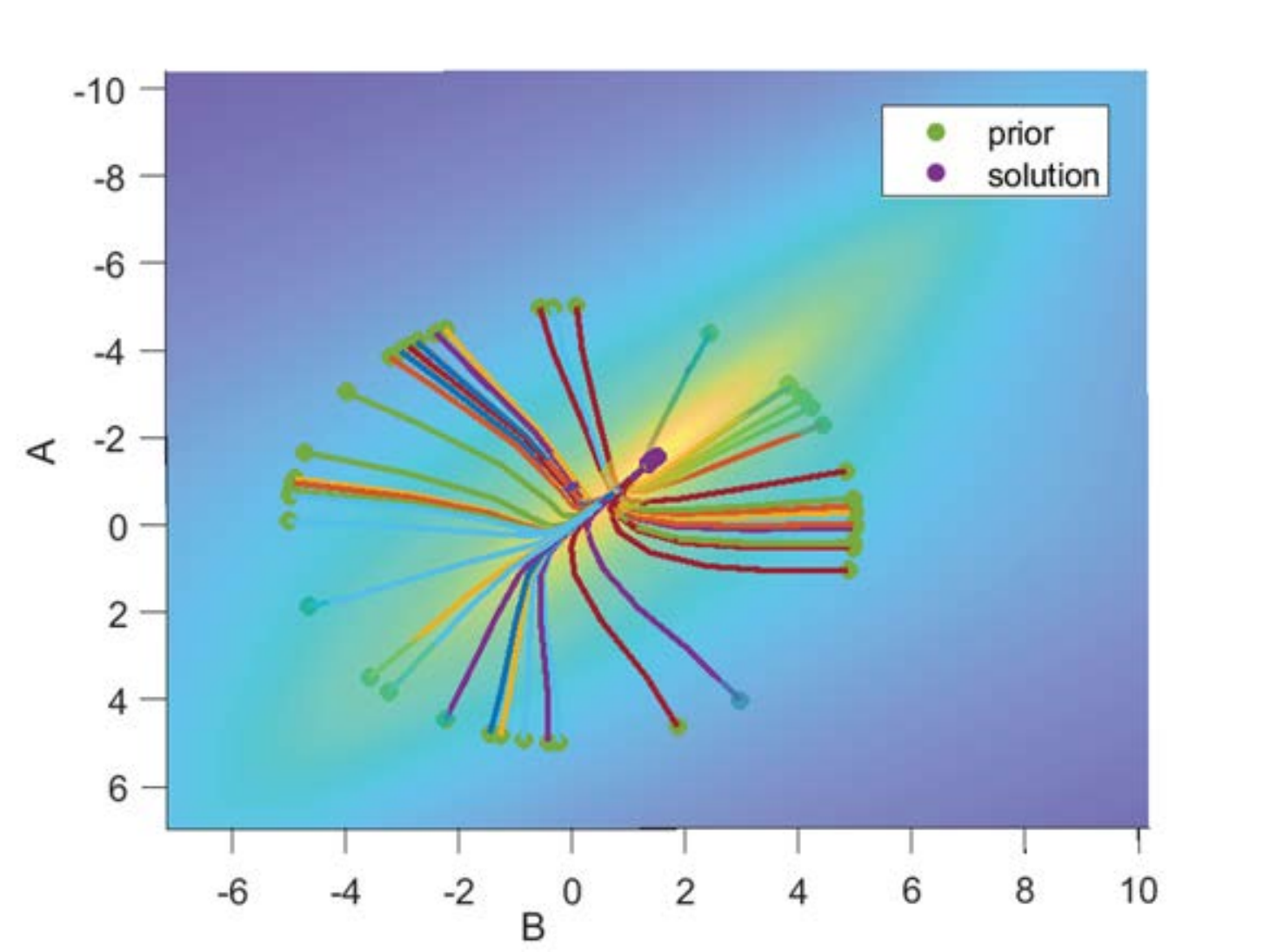}
		\caption{Top view of Fig. \ref{fig:climb_FE}. } \label{fig:climb_FE_top_view}
    \end{subfigure}   
    \caption{(a) The shape of the free energy manifold $\Bar{F}^o$ with respect to parameters $A$ and $B$ (both chosen as scalars for vizualization) changes with each E step iteration $i$, because of the interdependence between $\tilde{x}$ and $\theta$. Gradient ascend over $\Bar{F}^o$ at each E step sharpens the peak around the real parameters, where $\Bar{F}^o$ is the maximum. (b) Visualization of perception as a gradient ascend over free energy objective. 50 randomly sampled $\eta^\theta$ (green dots) lying on a circle climb up the free energy curve to converge to the same parameters (magenta dot) that coincides with the peak of one of the realizations of the free energy curve. (c) Top view of Fig. \ref{fig:climb_FE}. }\label{fig:FreeE_curves}
\end{figure*}

\subsection{Perception as Bayesian Inference}
The biological brain's perception is modelled as a Bayesian Inference which involves the computation of the posterior probability density $p(\vartheta/y)$ of parameter $\theta$, given the sensory measurement $y$ \cite{friston2008variational}. Since it involves the computation of an intractable integral $p(\vartheta/y) = {p(\vartheta,y)}/{\int p(\vartheta,y)d\vartheta}$, a variational density $q(\vartheta)$ called the recognition density is defined to closely approximate the posterior as $q(\vartheta)\approx p(\vartheta/y)$. This approximation is achieved by minimizing the Kullback-Leibler (KL) divergence of the distributions given by $KL(q(\vartheta)||p(\vartheta/y)) = \langle \ln q(\vartheta)\rangle_{q(\vartheta)} - \langle \ln{p(\vartheta/y)}\rangle_{q(\vartheta)}$, where $\langle.\rangle_{q(\vartheta)}$ represents the expectation over $q(\vartheta)$. Upon simplification using $p(\vartheta/y) = {p(\vartheta,y)}/{p(y)}$, it can be rewritten as: 
\begin{equation} \label{eqn:log_ev_def}
   \ln{p(y)} = \underbrace{ \langle \ln{p(\vartheta,y)}\rangle_{q(\vartheta)}  -\langle \ln q(\vartheta)\rangle_{q(\vartheta)}}_{\text{free energy}} +  KL(q(\vartheta)||p(\vartheta/y)),
\end{equation}
where $\ln p(y)$ is called the log-evidence. Since $\ln p(y)$ is independent of $\vartheta$, the  minimization of KL divergence for inference results in the maximization of free energy. This is the core idea behind using free energy as a proxy for perception through Bayesian Inference.

\subsection{Free energy objectives}
Two types of free energy objectives are used by DEM for perception: 1) free energy $F$ for the estimation of time varying components ($X = \begin{bmatrix} \tilde{x} \\  \tilde{v} \end{bmatrix}$) and 2) free energy action $\Bar{F} = \int F dt$ for the estimation of time invariant components ($\theta$ and $\lambda$). The free energy $F$ emerges from Bayesian statistics (Variational Inference) as an upper bound on surprise \cite{friston2010free}, and can be written as the sum of its internal energy $U$, mean field term $W$ and the entropy term $H$ as: 
\begin{equation} \label{eqn:free_E}
        F = U + W + H.
\end{equation} 
After Laplace approximation and mean-field approximation, $U, W$ and $H$ for an LTI system can be simplified as \cite{friston2008variational,anilmeera2021DEM_LTI}:
\begin{equation}
\begin{split}
    U  = & - \frac{1}{2} \tilde{\epsilon}^{T} \tilde{\Pi} \tilde{\epsilon} -\frac{1}{2} {\epsilon}^{\theta T}P^\theta {\epsilon}^\theta 
                  -\frac{1}{2} {\epsilon}^{\lambda T}P^\lambda {\epsilon}^\lambda \\
                  & + \frac{1}{2} \ln |\tilde{\Pi}| + \frac{1}{2} \ln |P^\theta| + \frac{1}{2} \ln |P^\lambda|    ,\\
    W = & \  tr(\Sigma^{\tilde{x}}U_{\tilde{x} \tilde{x}}+ \Sigma^{\tilde{v}}U_{\tilde{v} \tilde{v}} + \Sigma^\theta U_{\theta\theta} + \Sigma^\lambda U_{\lambda\lambda}),\\
    H = & \ \frac{1}{2} \ln |\Sigma^\theta| + \frac{1}{2} \ln |\Sigma^\lambda| + \frac{1}{2} \ln |\Sigma^{\tilde{x}}| + \frac{1}{2}\ln |{\Sigma}^{\tilde{v}}|,
\end{split}
\end{equation}
where $\tilde{\epsilon} = 
\begin{bmatrix}
\tilde{\textbf{y}} - \tilde{C}\tilde{x} \\ \tilde{v} - \tilde{\eta}^v \\ D^x\tilde{x} - \tilde{A}\tilde{x} - \tilde{B}\tilde{v}
\end{bmatrix}$, $\epsilon^\theta=\bm{\theta}-\theta$ and $\epsilon^\lambda =  \bm{\lambda}-\lambda $ are the prediction error for components in generalized coordinates, parameters and hyperparameters respectively. The prediction errors are precision weighed with the generalized precision matrix $\tilde{\Pi} = diag(\tilde{\Pi}^z,P^{\tilde{v}},\tilde{\Pi}^w)$, where $diag(.)$ is the block diagonal operation. Here $\Sigma^{\tilde{x}}$, $\Sigma^{\tilde{v}}$, $\Sigma^\theta$ and $\Sigma^\lambda$ are the covariance matrices denoting the uncertainty in the estimation of states, inputs, parameters and hyperparameters respectively. The free energy action $\Bar{F}$ can be written as \cite{anilmeera2021DEM_LTI}:
\begin{equation} \label{eqn:F_bar_MF}
\begin{split}
        \Bar{F} = & -\frac{1}{2} {\epsilon}^{\theta T}P^\theta {\epsilon}^\theta 
                  -\frac{1}{2} {\epsilon}^{\lambda T}P^\lambda {\epsilon}^\lambda +\frac{1}{2} \sum_{\vartheta^i} \sum_t   W^{\vartheta^i} \\
                  & +\frac{1}{2} \sum_t  \Big(- \tilde{\epsilon}^{T} \tilde{\Pi} \tilde{\epsilon} +  \ln |\tilde{\Pi}| +    \ln |\tilde{\Sigma}^x| +  \ln |\tilde{\Sigma}^v| \Big) \\
                  & + \frac{1}{2} \ln | P^\theta| + \frac{1}{2} \ln | P^\lambda|  + \frac{1}{2} \ln |\Sigma^\theta| + \frac{1}{2} \ln |\Sigma^\lambda| ,
\end{split}
\end{equation}
where $W^{\vartheta^i} = tr(\Sigma^{\vartheta^i}U_{\vartheta^i \vartheta^i})$ is the mean field term of $\vartheta^i \in \{\tilde{x}, \tilde{v}, \theta, \lambda\}$. $\Bar{F}$ can be seen as a generalized objective for Expectation Maximization (EM) algorithm with additional capabilities to handle colored noise. Removing generalized coordinates, brain's priors and the mean-field terms equates the objective functions of EM and DEM.

\begin{figure}[!b]
\centering
\captionsetup{justification=justified,margin=0cm}
\includegraphics[scale = 0.5]{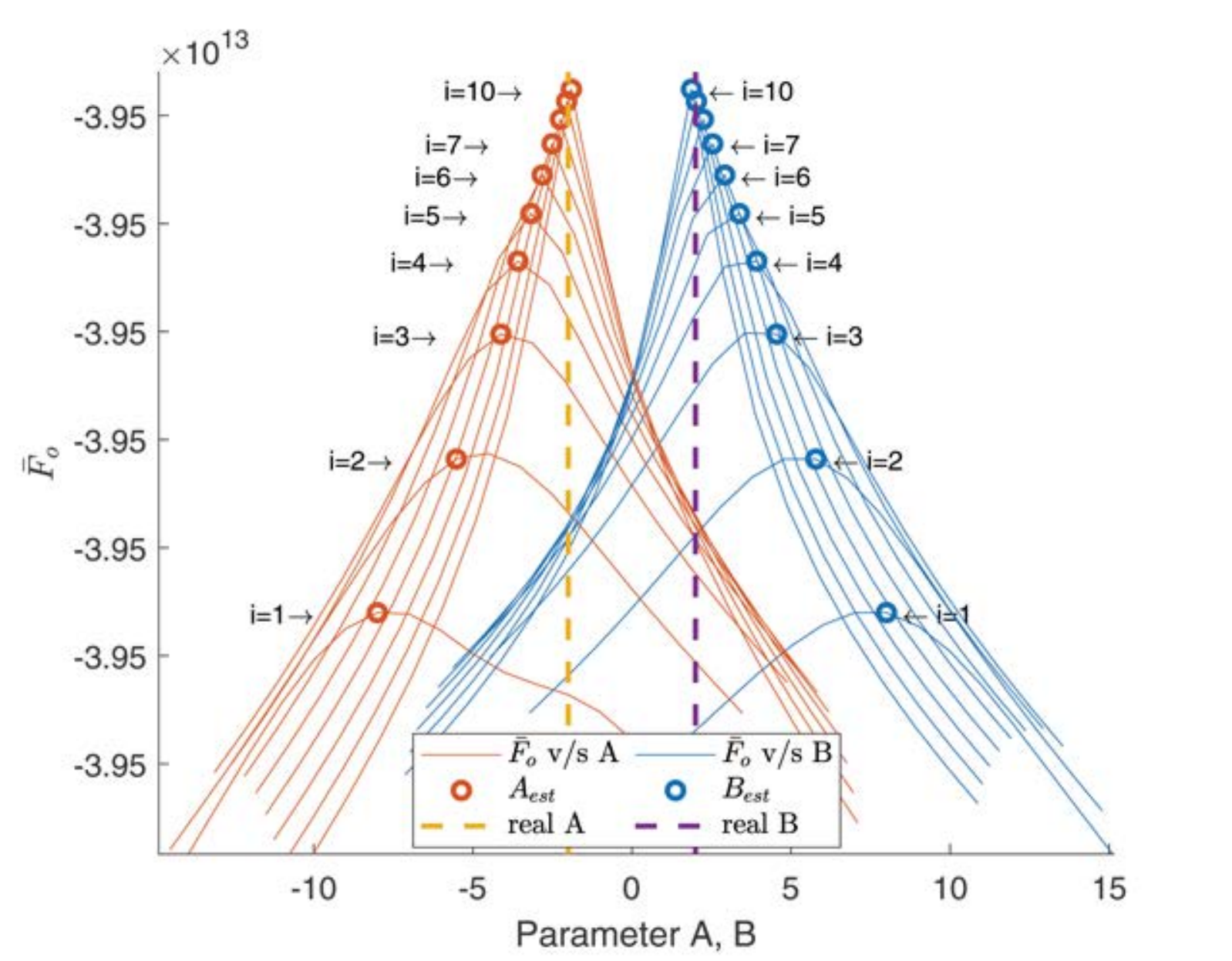}
\caption{Perception as the maximization of $\Bar{F}^o$. The parameter estimates (circles) start at $i=1$ with wrong priors ($\eta^A=-8,\eta^B=8$) and converges to the correct parameters (yellow and violet lines) at $A=-2$ and $B=2$. The $\Bar{F}^o$ curves (red and blue) are the cross sections of the manifold similar to the one in Fig. \ref{fig:f_bar} at different E step iteration $i$. The peak of these curves rise and narrows around the correct parameter until convergence. The increasing curvature of these peaks during learning is indicative of increasing confidence in estimation $\Pi^\theta$ as given in Equation \ref{eqn:Pi_curvature}. }
\label{fig:para_FE_opt}
\end{figure}

\subsection{Perception as free energy optimization}
DEM models the brain's inference process probabilistically through the estimation of two main components: the mean estimate and the uncertainty (inverse precision) in estimation. The mean estimate is computed through a gradient ascend on the free energy manifold. Accordingly, the update equation at time $t$, $a^{th}$ parameter update and $b^{th}$ hyperparameter update can be written as \cite{friston2008variational}:
\begin{equation} \label{eqn:GA_basic}
    \frac{\partial X}{\partial t} =  DX + k^X \frac{\partial F}{\partial X}, \ \  \frac{\partial \theta}{\partial a} =   k^\theta \frac{\partial \Bar{F}}{\partial \theta}, \ \ \frac{\partial \lambda}{\partial b} =   k^\lambda \frac{\partial \Bar{F}}{\partial \lambda},
\end{equation}
where $k^X,k^\theta$ and $k^\lambda$ are the learning rates. $\Bar{F}$ is maximized with respect to the estimation uncertainty $\Sigma^{\vartheta^i}$ when the first gradient is zero and the second gradient is negative definite. 
\begin{equation} \label{eqn:grad_ascend}
\begin{split}
        \frac{\partial \Bar{F}}{\partial \Sigma^{\vartheta^i}} = & \frac{1}{2} \frac{\partial }{\partial \Sigma^{\vartheta^i}} \Big(  \ln |\Sigma^{\vartheta^i}| +  \sum_t tr(\Sigma^{\vartheta^i}U_{\vartheta^i \vartheta^i}) \Big) \\
        =  & \frac{1}{2} \big( (\Sigma^{\vartheta^i})^{-1} + \Bar{U}_{\vartheta^i \vartheta^i} \big), \\
          \frac{\partial^2 \Bar{F}}{(\partial \Sigma^{\vartheta^i})^2}  = & -\frac{1}{2} (\Sigma^{\vartheta^i})^{-2} \prec O.
\end{split}
\end{equation}
Forcing the first gradient to zero yields the optimal precision (inverse covariance) of estimates as:
\begin{equation} \label{eqn:Pi_curvature}
         {\Pi}^{\tilde{x}} = -U_{\tilde{x}\tilde{x}}, \ {\Pi}^{\tilde{v}} = -U_{\tilde{v}\tilde{v}}, \  \Pi^\theta = - \Bar{U}_{\theta\theta}, \ \Pi^\lambda = - \Bar{U}_{\lambda\lambda}
\end{equation}

Note that $\Bar{U}$ and $U$ are used for time independent and time dependent $\vartheta^i$ respectively. Therefore, the mean estimates and the uncertainty in their estimation can be obtained from Equations \ref{eqn:GA_basic} and \ref{eqn:Pi_curvature}, only by using the first two gradients of the energy terms ($F,\Bar{F},U,\Bar{U}$). Substituting Equation \ref{eqn:Pi_curvature} in \ref{eqn:F_bar_MF} eliminates the mean field terms and simplifies $\Bar{F}$ for an LTI system at optimal precision as the sum of weighted prediction errors and entropy \cite{anilmeera2021DEM_LTI}:
\begin{equation} \label{eqn:F_action_opt}
    \begin{split}
        &\Bar{F}^o =   \frac{1}{2} n^t \Big[ \underbrace{ \ln |\tilde{\Pi}^z| +  \ln |\tilde{P}^v| +  \ln |\tilde{\Pi}^w| }_{\text{noise entropy}} \Big] \\
                  & +\underbrace{\frac{1}{2} n_t \ln |\tilde{\Sigma}^X|}_{\text{state and input entropy}}  
                    +\underbrace{\frac{1}{2} \ln |\Sigma^\theta P^\theta|}_{\text{parameter entropy}}
                    +\underbrace{\frac{1}{2} \ln |\Sigma^\lambda P^\lambda|}_{\text{hyperparameter entropy}} \\ 
                  & - \frac{1}{2} \sum_t \Big[ \underbrace{ (\tilde{\textbf{y}}-\tilde{C}\tilde{x})^T \tilde{\Pi}^z (\tilde{\textbf{y}}-\tilde{C}\tilde{x}) }_{\text{prediction error of outputs}}  + \underbrace{ (\tilde{v}-\tilde{\eta}^v)^T \tilde{P}^v (\tilde{v}-\tilde{\eta}^v) }_{\text{prediction error of inputs}} \Big] \\
        &  - \frac{1}{2} \sum_t \Big[ \underbrace{  (D^x\tilde{x}-\tilde{A}\tilde{x}-\tilde{B}\tilde{v})^T \tilde{\Pi}^w  (D^x\tilde{x}-\tilde{A}\tilde{x}-\tilde{B}\tilde{v}) }_{\text{prediction error of states}}  \Big]   \\
                  & - \frac{1}{2} \underbrace{ (\theta-\eta^\theta)^T P^\theta (\theta-\eta^\theta)}_{\text{prediction error of parameters}}
                    - \frac{1}{2} \underbrace{ (\lambda-\eta^\lambda)^T P^\lambda (\lambda-\eta^\lambda)}_{\text{prediction error of hyperparameters}} \\
    \end{split}
\end{equation}
Maximizing $\Bar{F}$ for perception is equivalent to minimizing the prediction error, while maximizing the uncertainty in estimation through the entropy terms. This acts like regularization, preventing the brain from overfitting the model to the data, making it an ideal objective function for robot learning.

\begin{table*}[!b]
 \caption{LINEARIZED QUADCOPTER MODELS of DIFFERENT ORDER.}
 \label{tab:systems}
\centering
\setlength{\tabcolsep}{6pt}
\begin{tabularx}{\textwidth}{ccccccc} 
 \toprule
             & Order & x & A & B & C & v \\
 \toprule
 System 1  & 1  & $\dot{\phi}$ & 0 & $\begin{bmatrix} 0.3748 &  -0.3748 &  -0.3748 & 0.3748 \end{bmatrix}$ & $I_1$ & $\begin{bmatrix}
    pwm^1 & pwm^2 & pwm^3 & pwm^4 
\end{bmatrix}^T$  \\
 \midrule
 System 2  & 2  & $\begin{bmatrix} \phi \\ \dot{\phi} \end{bmatrix}$ & $\begin{bmatrix} 0 & 1 \\ 0 & 0 \end{bmatrix}$ & $\begin{bmatrix} 0 & 0& 0 & 0 \\ 0.3748 &  -0.3748 &  -0.3748 & 0.3748 \end{bmatrix}$ & $I_2$ & $ \begin{bmatrix}
    pwm^1 & pwm^2 & pwm^3 & pwm^4 
\end{bmatrix}^T$ \\
 \midrule
 System 3  & 3  & $\begin{bmatrix} \dot{y} \\ \phi \\ \dot{\phi} \end{bmatrix}$ & $\begin{bmatrix} 0 &  -9.81   &   0 \\ 0 & 0 & 1 \\ 0 & 0 & 0 \end{bmatrix}$ & $\begin{bmatrix} 0 & 0& 0 & 0 \\ 0 & 0& 0 & 0 \\ 0.3748 &  -0.3748 &  -0.3748 & 0.3748 \end{bmatrix}$ & $I_3$ & $\begin{bmatrix}
    pwm^1 & pwm^2 & pwm^3 & pwm^4 
\end{bmatrix}^T$ \\
 \midrule
 System 4  & 4  & $\begin{bmatrix} y \\ \dot{y} \\ \phi \\ \dot{\phi} \end{bmatrix}$ & $\begin{bmatrix} 0 & 1 & 0 & 0 \\ 0&  0 &  -9.81   &   0 \\ 0& 0 & 0 & 1 \\ 0& 0 & 0 & 0 \end{bmatrix}$ & $\begin{bmatrix} 0 & 0 & 0 & 0 \\ 0 & 0& 0 & 0 \\ 0 & 0& 0 & 0 \\ 0.3748 &  -0.3748 &  -0.3748 & 0.3748 \end{bmatrix}$ & $I_4$ & $\begin{bmatrix}
    pwm^1 & pwm^2 & pwm^3 & pwm^4 
\end{bmatrix}^T$ \\
 \bottomrule
\end{tabularx}
\end{table*}

\subsection{Dynamic Expectation Maximization}
DEM postulates the brain's perception as a gradient ascend of its free energy objectives using three steps \cite{friston2008variational}:
\begin{itemize}
    \item D step - state ($\tilde{x}$) and input ($\tilde{v}$) estimation,
    \item E step - parameter ($\theta$) estimation and
    \item M step - hyperparameter ($\lambda$) estimation.
\end{itemize}
DEM results in Gaussian probability distributions with its mean as the estimate and its standard deviation as the uncertainty in estimation. The D steps follows the gradient ascend over $F$ given in Equation \ref{eqn:free_E} to estimate $\tilde{x},\tilde{v},\Pi^{\tilde{x}}$ and $\Pi^{\tilde{v}}$. The E and M steps follows the gradient ascend over $\Bar{F}$ given in Equation \ref{eqn:F_bar_MF} to estimate $\theta,\Pi^\theta$ and $\lambda,\Pi^\lambda$ respectively. We use the reformulated version of DEM for LTI systems from our previous work \cite{anilmeera2021DEM_LTI} for rest of the paper. Fig. \ref{fig:FreeE_curves} demonstrates DEM's parameter learning procedure, whereas Fig. \ref{fig:para_FE_opt} demonstrates the evolution of the cross sections of $\Bar{F}^o$ manifold during perception.

\begin{figure}[!htb]
\centering
\captionsetup{justification=centerlast,margin=0cm}
\includegraphics[scale = 0.15]{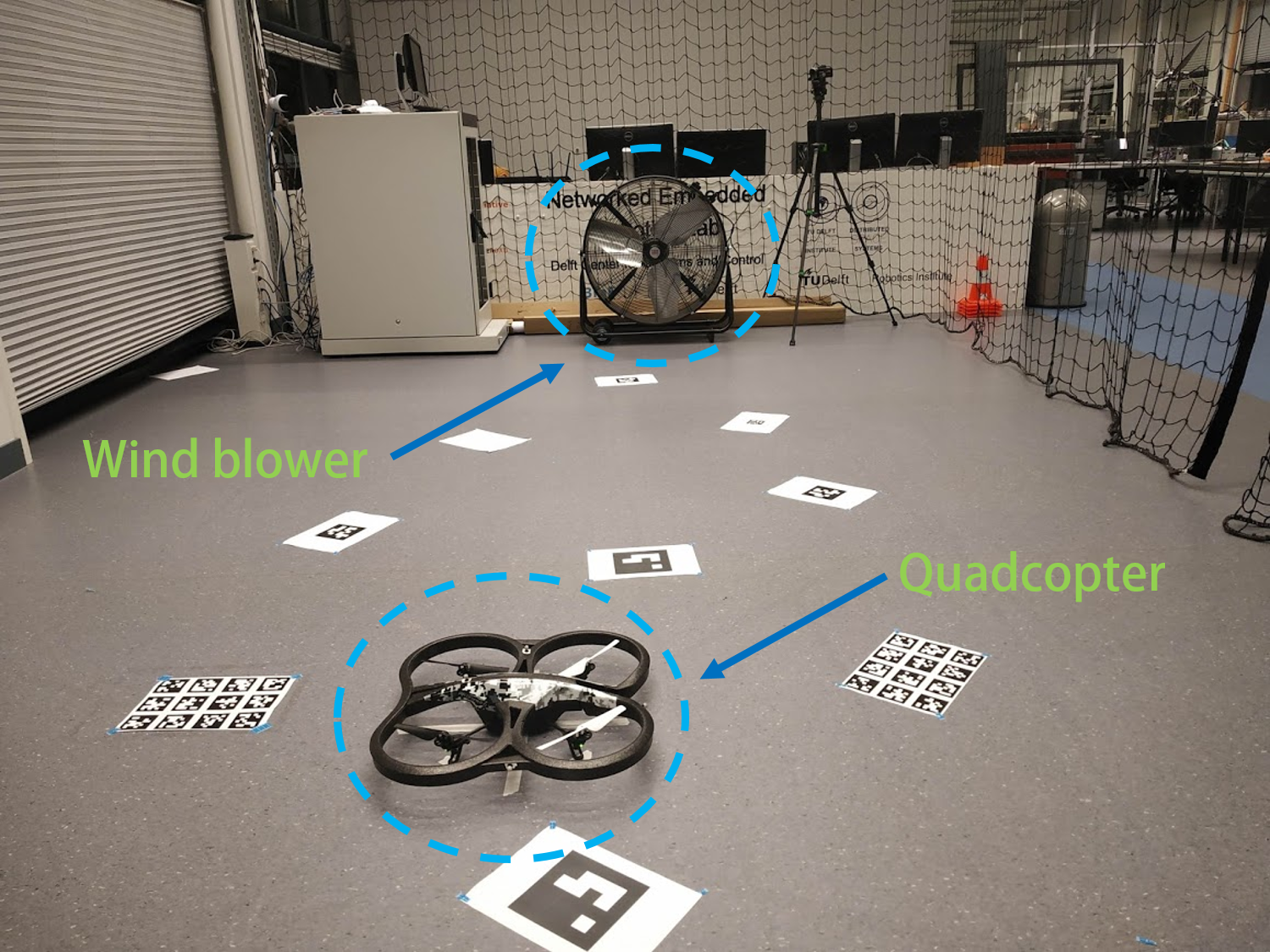}
\caption{The quadcopter and the wind blower in the lab environment.}
\label{fig:drone}
\end{figure}

\section{EXPERIMENTAL RESULTS AND ANALYSIS}
This section aims to provide a proof of concept for DEM through real quadrotor flights in wind conditions.

\subsection{Experimental setup}
A Parrot AR drone 2.0 was used to hover in wind in a controlled lab environment as shown in Fig. \ref{fig:drone}. A simple linear state space model (Equation \ref{eqn:general_LTI}) that maps four rotor PWM signals of the quadrotor to its roll angle $\phi$ and roll angular velocity $\dot{\phi}$ is constructed (System 2 in Table \ref{tab:systems}). We consider a simple system where the optitrack motion capture system directly observes the internal states ($x$) of the quadcopter through measurements $y$ with a precision of measurement noise $\Pi^z$. The model is derived from \cite{benders2020ar} after linearization around the equilibrium point. The key idea behind this experimental design is to introduce the linearization error and the error from unmodelled wind dynamics to the system as colored process noise $w$.

\begin{figure}[!htb]
\centering
\captionsetup{justification=justified,margin=0cm}
\includegraphics[scale=.8,trim={1cm 0 0.3cm 0},clip]{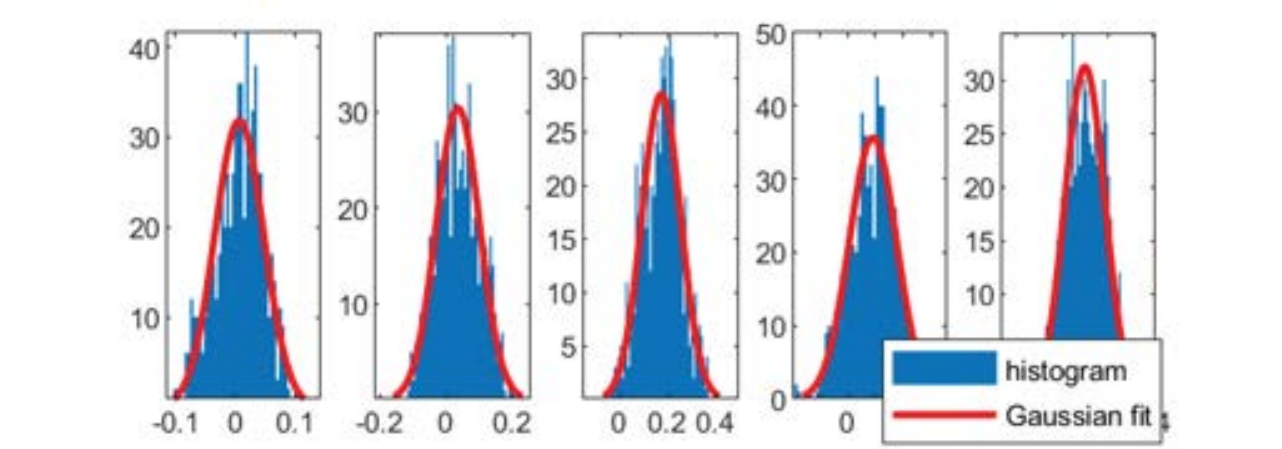}
\caption{The histograms of process noise $w^{\dot{\phi}}$ for all five flight experiments follow a Gaussian distribution. }
\label{fig:histograms}
\end{figure}

\subsection{Data preparation}
We consider five quadrotor flights under different wind conditions (wind speed and blower orientation), all for a duration of 850 time steps each with $dt = 0.0083s$. Although different wind conditions might induce different levels of noise color, it doesn't influence our final result. We split each time series data into two parts: training data (700 time steps $\simeq$ 80$\%$) and test data (150 time steps $\simeq$ 20$\%$). The training data is used to learn the model, whereas the test data (unseen data) is used to test the performance of the learned model for output prediction. As a pre-processing step, the input pwm signal to the rotor was mean shifted to zero, and scaled down from high values using: $ v = \frac{v-mean(v)}{max(v)-min(v)} $.

\begin{figure*}[!t]
    \centering
    \begin{subfigure}[b]{0.32\textwidth}
		\includegraphics[width=\textwidth,trim={0.3cm 0 0.3cm 0},clip]{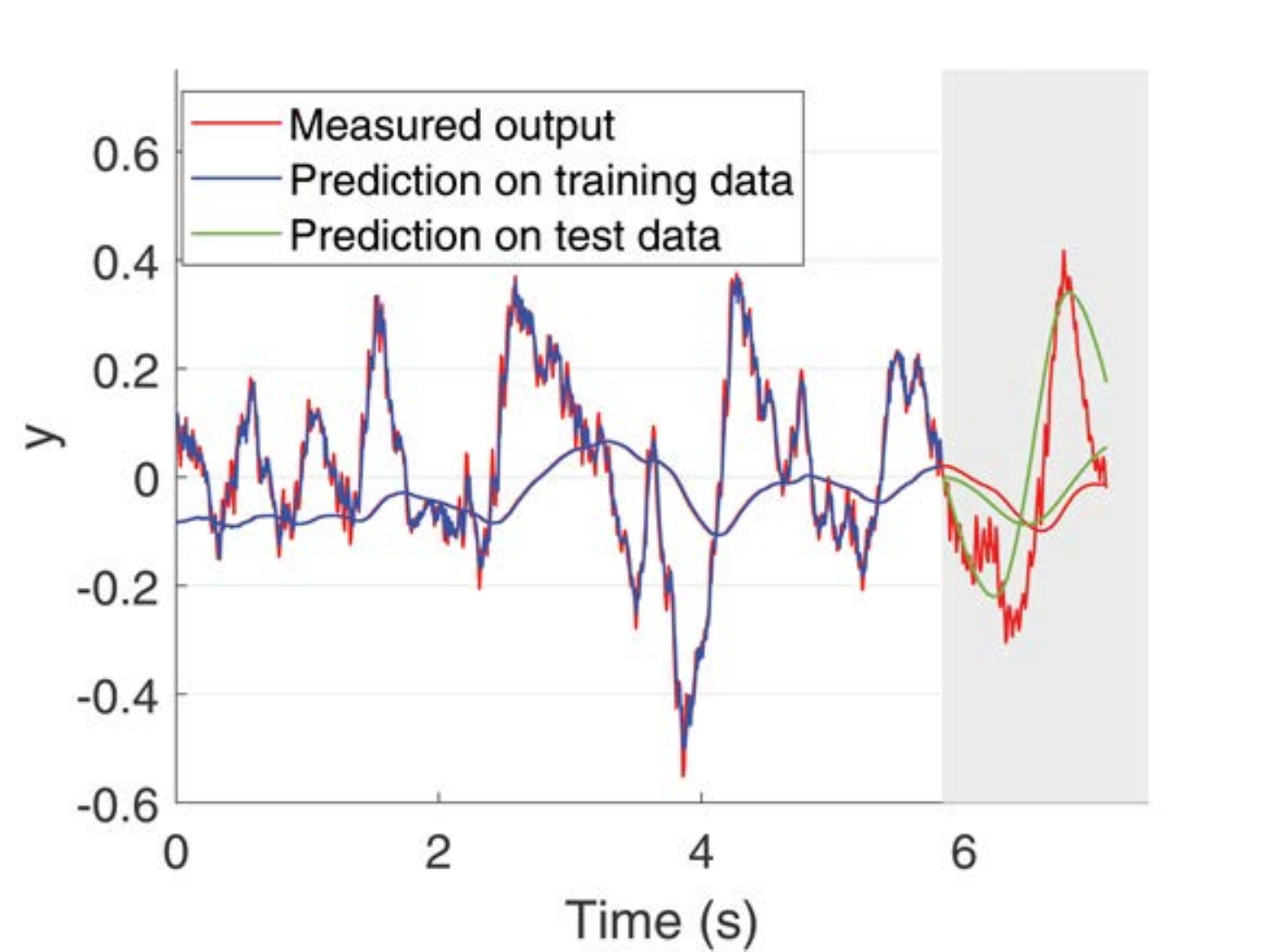}
		\caption{Output prediction.}\label{fig:y_pred_DEM}
    \end{subfigure} 
    \begin{subfigure}[b]{0.32\textwidth}
		\includegraphics[width=\textwidth,trim={.2cm 0 .3cm 0},clip]{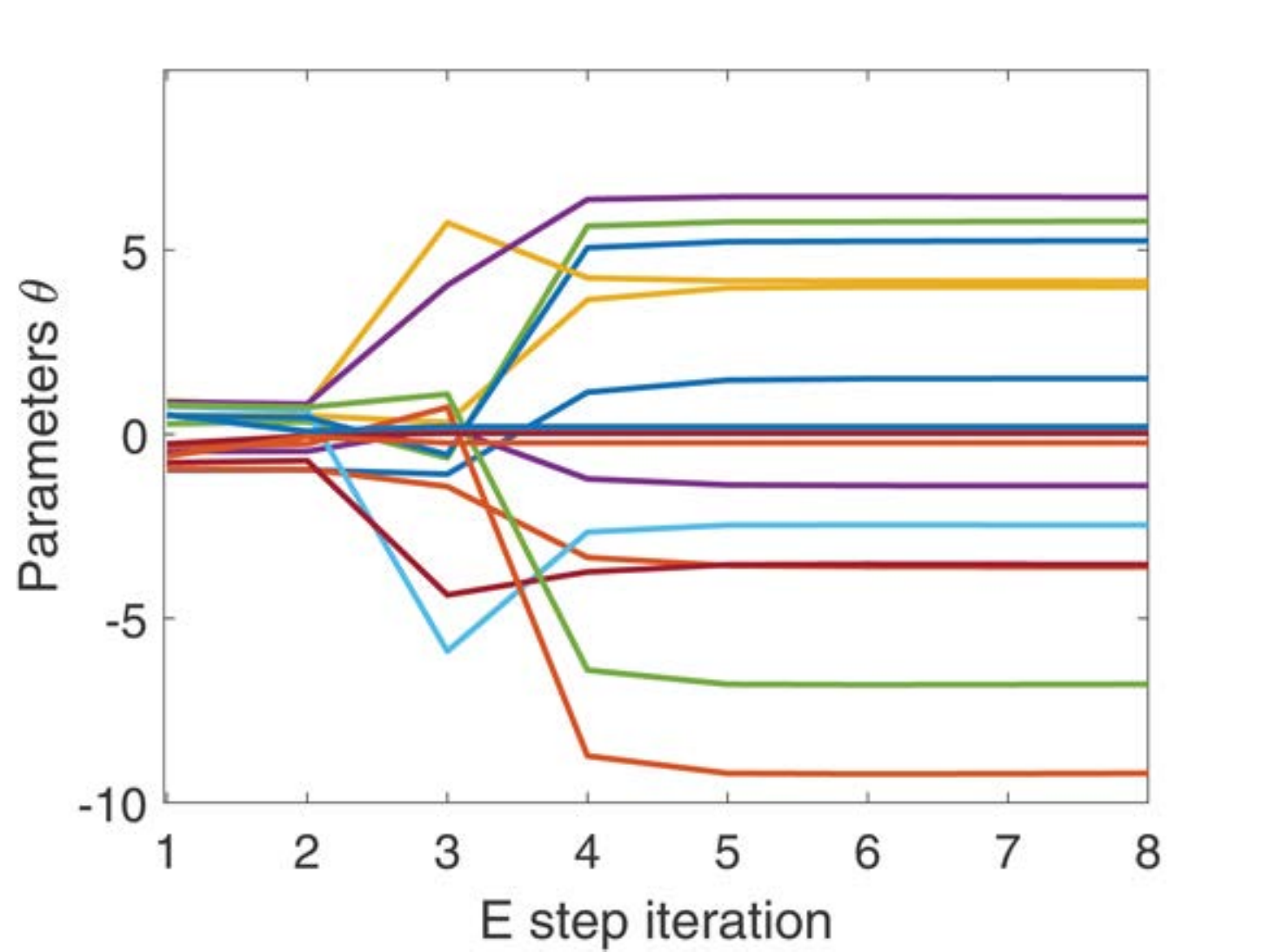}
		\caption{Parameter estimation. }\label{fig:parameter_learning_sample}
    \end{subfigure}    
    \begin{subfigure}[b]{0.3\textwidth}
		\includegraphics[width=\textwidth,trim={.2cm 0cm .3cm 0},clip]{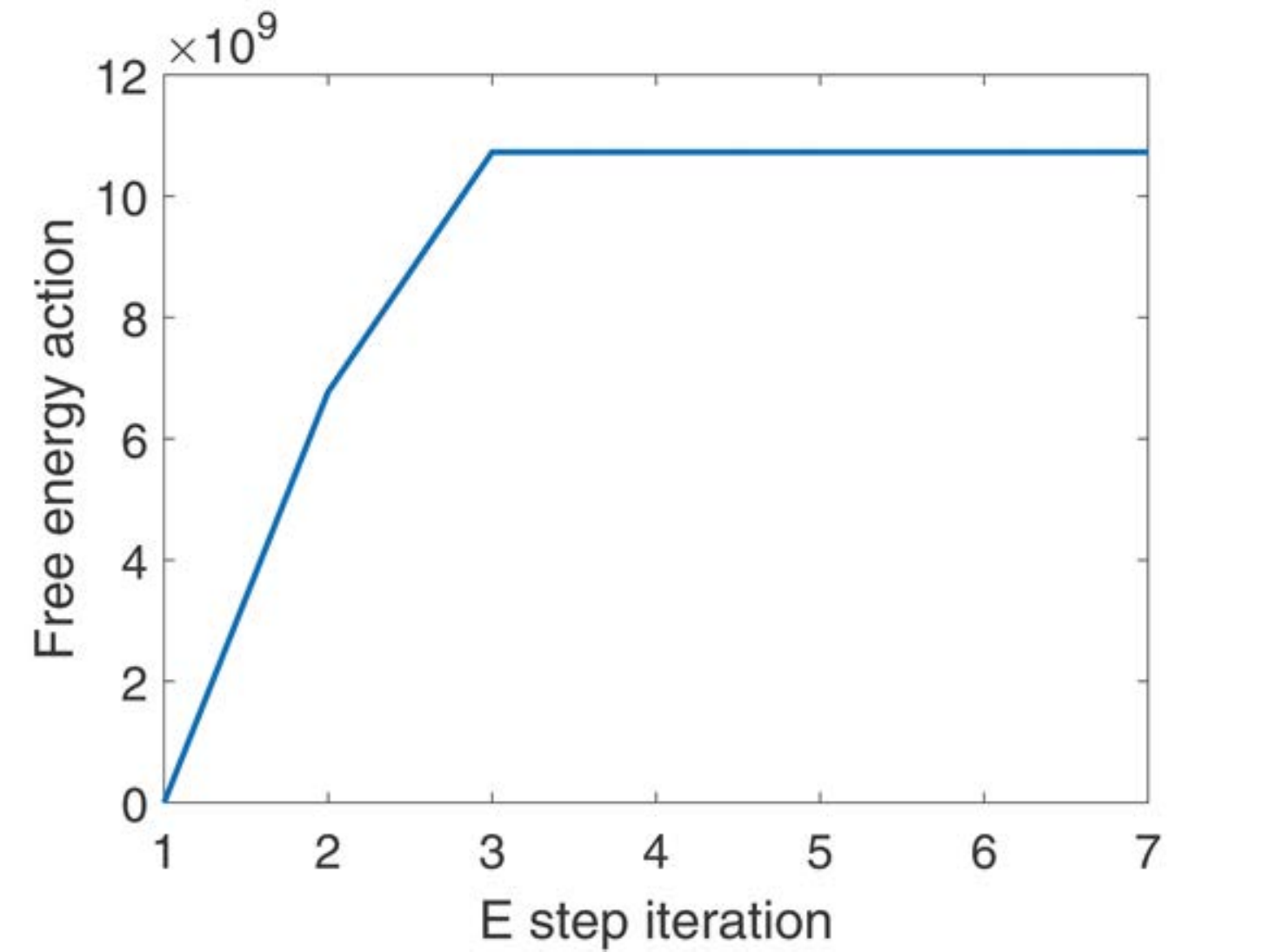}
		\caption{Maximize free energy action $\Bar{F}(i)-\Bar{F}(0)$. }\label{fig:FreeE_sample}
    \end{subfigure}   
    \caption{The robot brain's perception as a free energy optimization scheme (DEM) for the output prediction of a quadcopter hovering under wind conditions. (a) The coinciding blue and red curves demonstrate that DEM can accurately perform one step ahead output prediction on the training data (white background). The green curve following the trend of the red curve demonstrates that DEM can perform 150 step ahead output predictions on unseen data (grey background) using the learned model. (b) The parameter estimation step (E step) explores the parameter space to finally converge to a solution for $A$, $B$ and $C$ matrices. (c) Perception driven by the maximization of $\Bar{F}$.}\label{fig:DEM_estimation_eg}
\end{figure*}

\begin{figure}[!b]
\centering
\includegraphics[scale=.4,trim={.3cm 0 1.1cm 0},clip]{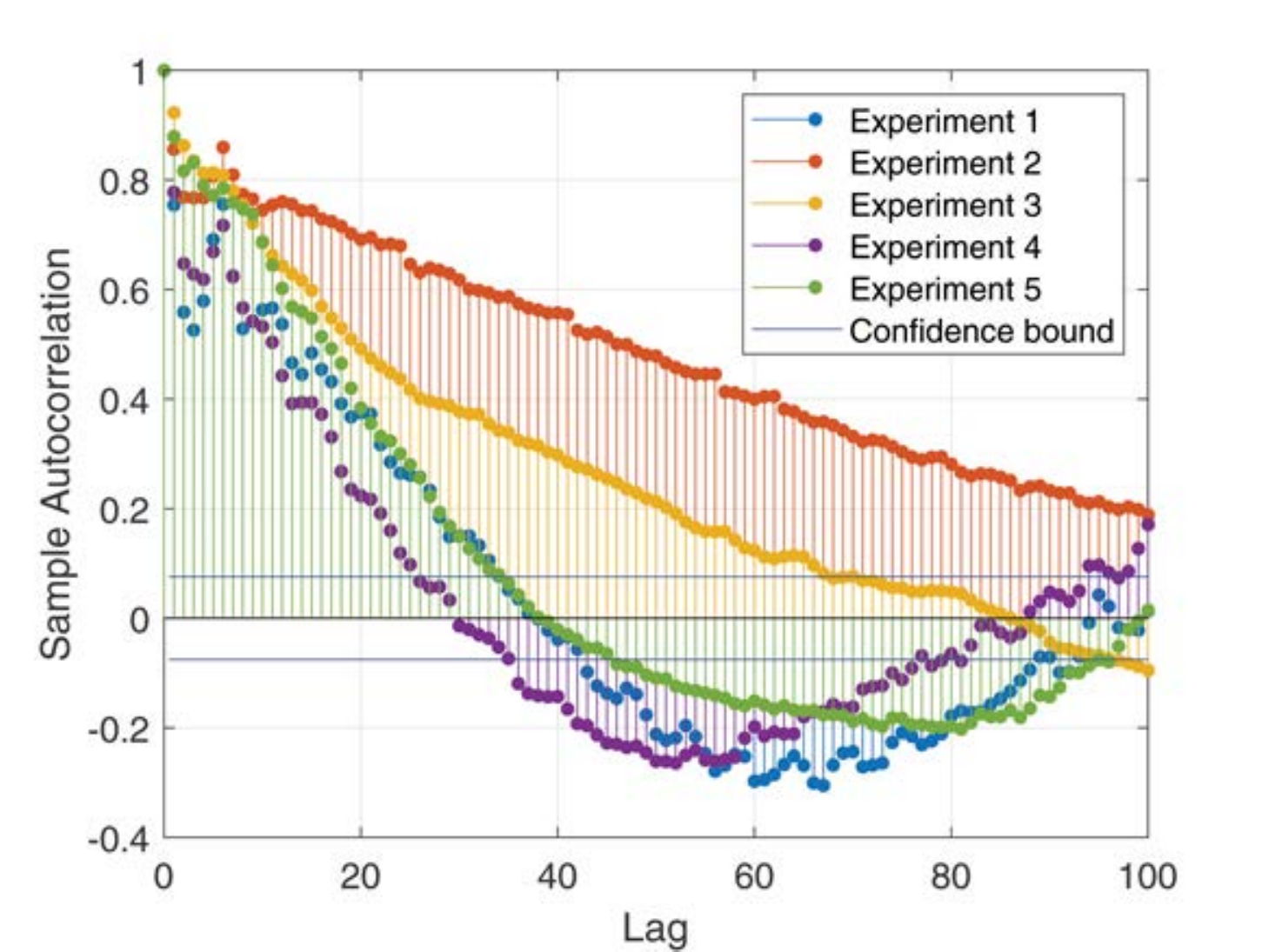}
\caption{The autocorrelation plot of process noise ($w^{\dot{\phi}}$) for all five experiments doesn't drop within the confidence bound immediately after zero lag, confirming the presence of a range of noise color. }
\label{fig:autocorr}
\end{figure}

\subsection{Noise color and Laplace approximation}
We validate the two fundamental assumptions of DEM, the Laplace approximation and the noise color assumption, by using noise histogram and noise autocorrelation graph respectively on all experiments. Fig. \ref{fig:histograms} demonstrates that the histogram of process noise $w^{\dot{\phi}}$ is Gaussian in nature, confirming DEM's Laplace approximation. Fig. \ref{fig:autocorr} demonstrates that the autocorrelation plot of $w^{\dot{\phi}}$ does not correspond to white noise where it should have been bounded within the confidence bounds for lags above 0. This confirms the presence of strong color in process noise, which was induced by unmodelled wind dynamics and linearization errors.

\subsection{Algorithm settings for DEM} \label{sec:alg_settings}
The parameter priors $\eta^\theta$ were randomly selected from [-1,1], and a moderate level of parameter precision ($P^\theta = e^4$) was set to encourage exploration in the parameter space, starting from random priors $\eta^\theta$. A high observation noise hyperparameter ($\lambda^z=20$) was used with high confidence ($P^{\lambda^z}=e^{25}$) to represent the accurate motion capture system measurements (optitrack). A low process noise hyperparameter ($\lambda^w=3$) was used with high confidence ($P^{\lambda^w}=e^{20}$) to represent high process noise emerging from wind and non-linearity errors. The noises were assumed to have a Gaussian temporal correlation with a noise smoothness of $s = dt$ for all the experiments. To handle colored noise, the generalized coordinate was used with an order of generalization for states and inputs as $p=2$ and $d=1$ respectively. DEM used the same settings to process all data.

\subsection{Output prediction using DEM} \label{sec:Output_prediction}
The robot brain's perception of a quadrotor hovering in wind was emulated using the DEM algorithm. The quadrotor model was learned by maximizing $\Bar{F}$ using the experiment 2 data and the result is shown in Fig. \ref{fig:DEM_estimation_eg}. The parameter estimation (E step) explores the parameter space and converges to a solution within few iterations (Fig. \ref{fig:parameter_learning_sample}), despite starting from wrong random priors ($\eta^\theta$) in the range [-1,1]. The learned model is tested for one step ahead output prediction on the training data, and for output predictions until 150 step ahead on the test data. The coinciding predictions and measured output in Fig. \ref{fig:y_pred_DEM} demonstrates DEM's successful model learning for output prediction, both on seen and unseen data. Fig. \ref{fig:FreeE_sample} shows the maximization of $\Bar{F}$ during perception.

\subsection{Metric for comparison}
We measure the quality of output prediction using the Mean Squared Prediction Error (MSPE) for 150 step ahead predictions on unseen test data:
\begin{equation}
     MSPE = \frac{1}{150} \sum_{i=T+1}^{T+150} ({y_{i}} - \hat{y}_{i})^2,
\end{equation}
where $y_i$ is the measured output and $\hat{y}_i$ is the output prediction at time step $i$. A high quality perception algorithm will have the least MSPE when compared to other methods. 
\begin{figure}[!b]
\centering
\includegraphics[scale=.4,trim={0.5cm 0 1cm 0},clip]{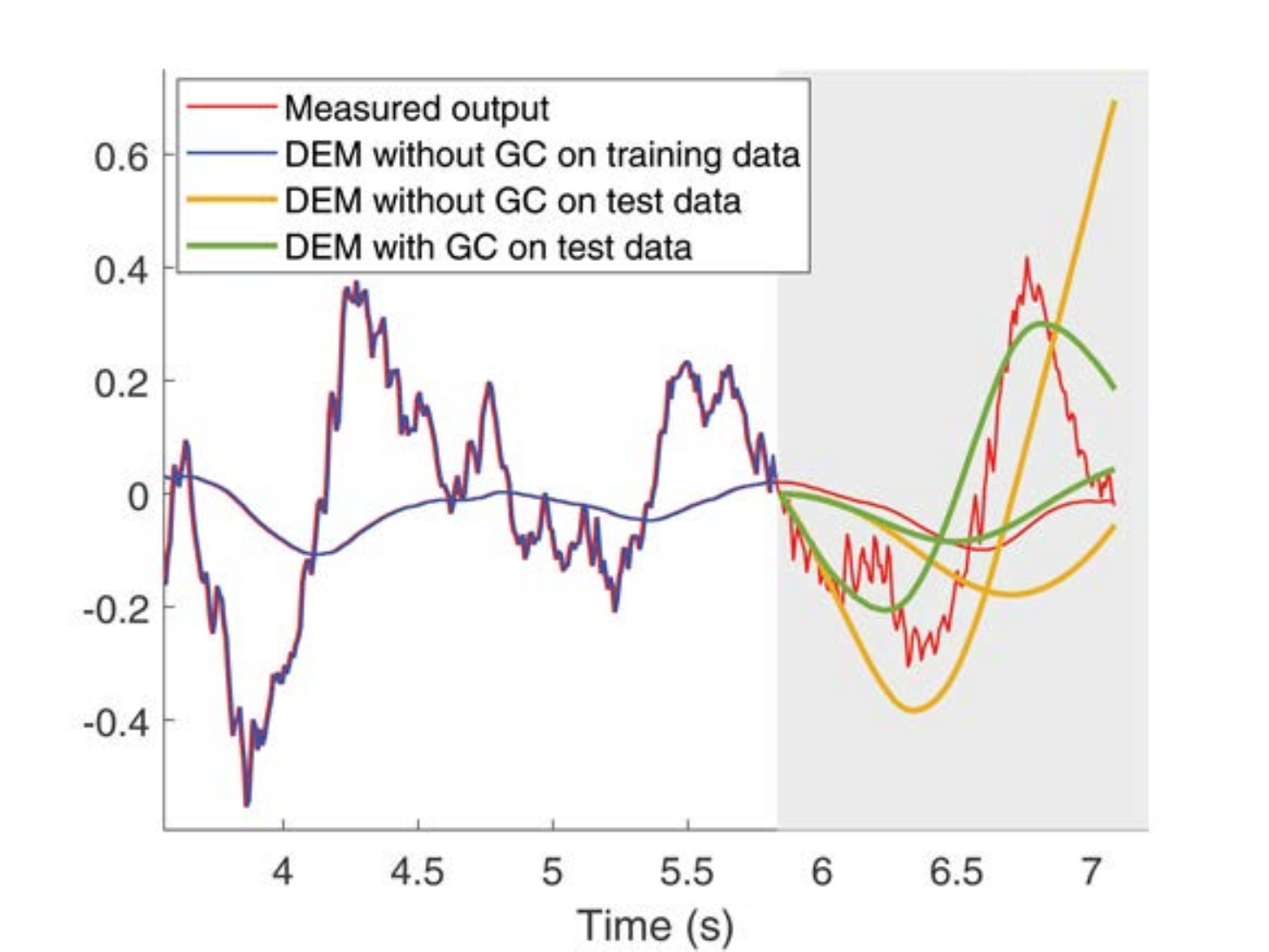}
\caption{The output prediction of DEM on the test data improves when GC is used during perception. The green curve follows the trend of the red curve better than the yellow curve.}
\label{fig:generalized_coor_y_pred}
\end{figure}

\begin{figure*}[!ht]
    \centering
    \begin{subfigure}[b]{0.32\textwidth}
		\includegraphics[width=\textwidth,trim={0.3cm 0 0.3cm 0},clip]{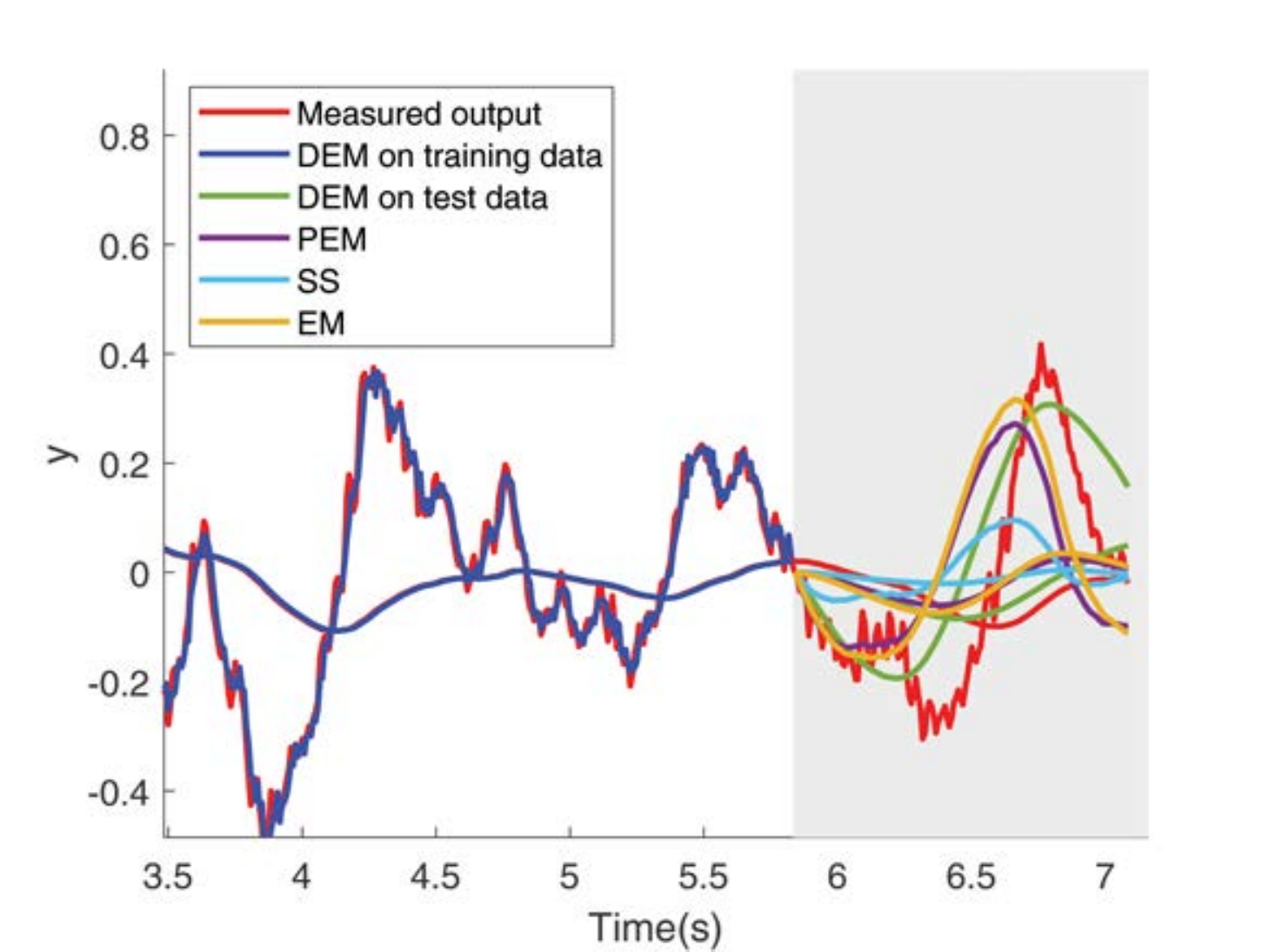}
		\caption{Benchmarking output prediction.}\label{fig:y_pred_benchmark}
    \end{subfigure} 
    \begin{subfigure}[b]{0.32\textwidth}
		\includegraphics[width=\textwidth,trim={.2cm 0 .3cm 0},clip]{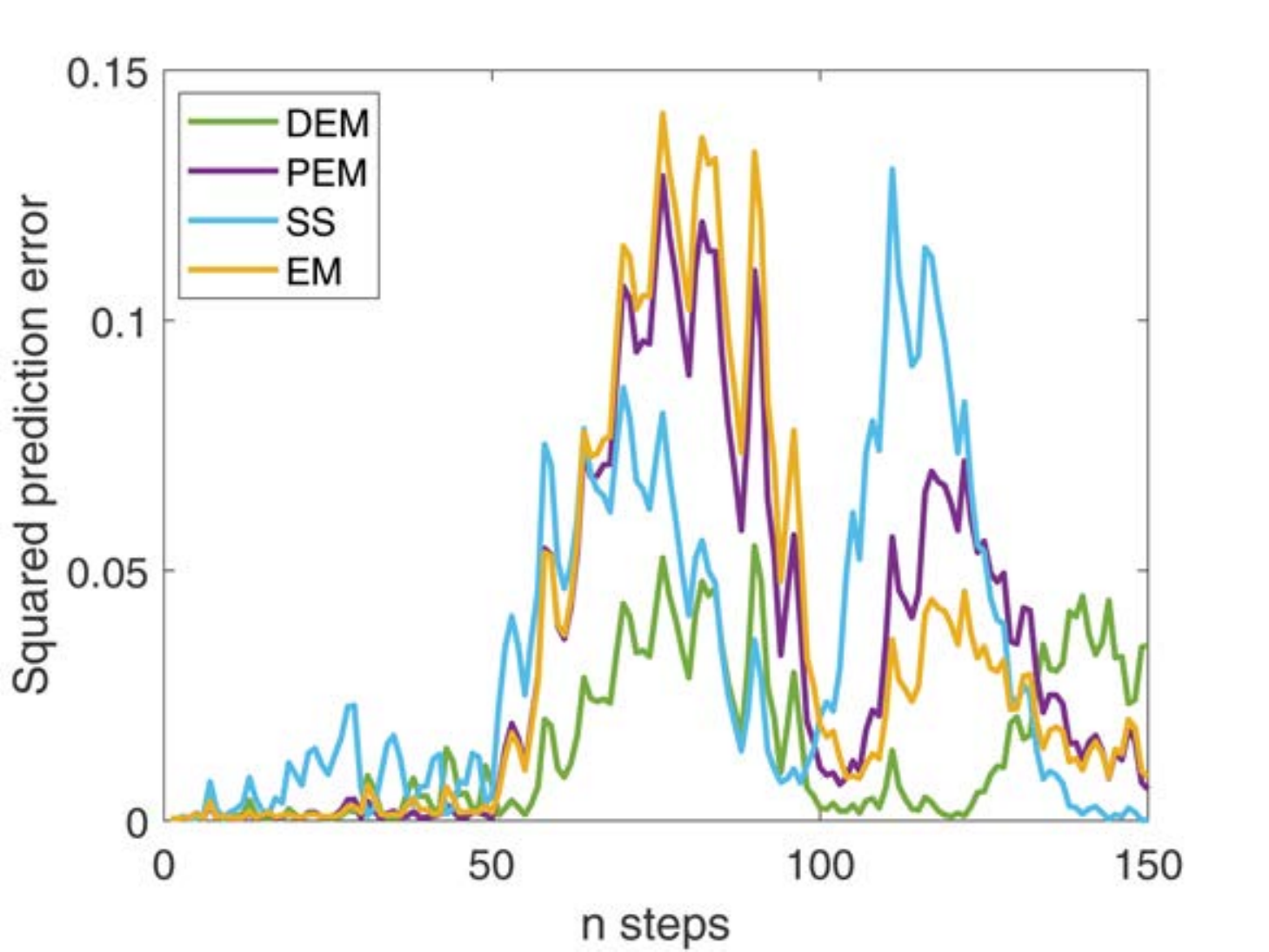}
		\caption{The n-step ahead prediction error. }\label{fig:pred_error}
    \end{subfigure}    
    \begin{subfigure}[b]{0.32\textwidth}
		\includegraphics[width=\textwidth,trim={.2cm 0 .3cm 0},clip]{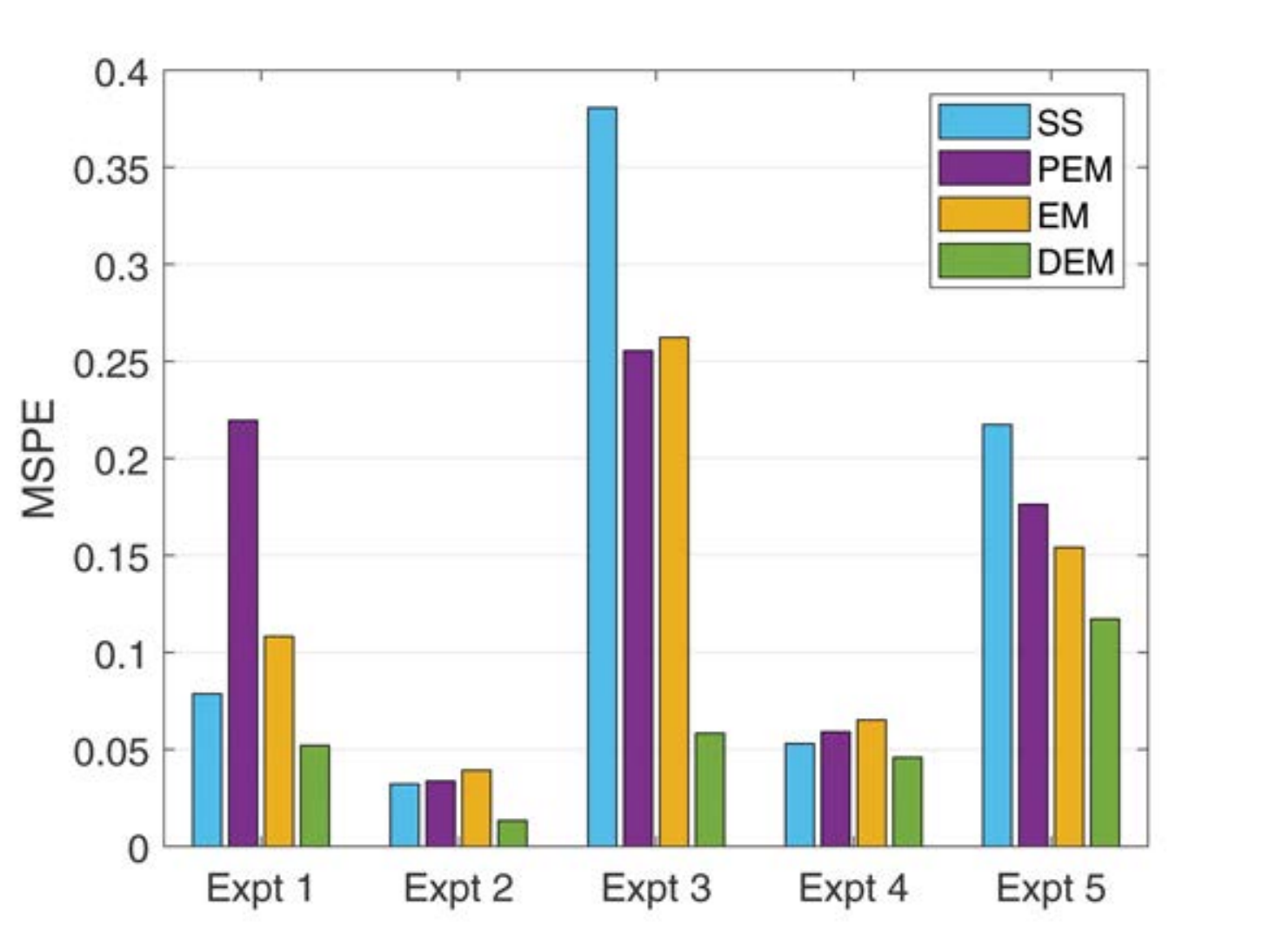}
		\caption{MSPE for five quadrotor experiments. }\label{fig:MSPE_bench}
    \end{subfigure}   
    \caption{The parameter learning using DEM is benchmarked against other system identification methods - PEM, SS and EM. (a) The output predictions of DEM (green curve) for experiment 2 on the unseen test data (grey background) best follows the trend of the measured output (red curve) when compared to other methods. (b) The corresponding n-step ahead output prediction error is the lowest for DEM (green curve) for most cases. (c) DEM outperforms other methods with the best quality output prediction by minimizing MSPE for all five experiments, thereby demonstrating that DEM is a very competitive algorithm.}\label{fig:MSPE_benchmarking}
\end{figure*}

\subsection{Importance of generalized coordinates} \label{sec:GC_imp}
The key difference between DEM and other classical estimators is its capability to deal with colored noise using the generalized coordinates (GC). In this section, we show that the use of generalized coordinates improves the accuracy of output prediction of a quadrotor flying in wind. We repeat the same procedure in Section \ref{sec:Output_prediction} for two different conditions: 1) output prediction with GC ($p=2$) and 2) without GC ($p=1$). Fig. \ref{fig:generalized_coor_y_pred} demonstrates that the use of GC provides a better output prediction than when no GC was used. MSPE was used to measure the quality of output prediction for all five flight experiments for both conditions (with and without GC), and the results are tabulated in Table \ref{tab:SSE_GC}. The results show a lower MSPE for DEM with GC when compared to DEM without using GC, revealing the importance of using GC for output prediction.

\begin{table}[h]
 \caption{INFLUENCE of GC on MSPE.}
 \label{tab:SSE_GC}
\centering
\setlength{\tabcolsep}{4pt}
\begin{tabularx}{\columnwidth}{ccccccc} 
 \toprule
             & expt 1 & expt 2 & expt 3 & expt 4 & expt 5 & total \\
 \midrule
 DEM without GC  & 0.1197 & 0.0640 & 0.1647 & 0.0518 & 0.0951 & 0.4953 \\
 DEM with GC     & 0.0521 & 0.0133 & 0.0583 & 0.0458 & 0.1172 & \textbf{0.2867} \\
 \bottomrule
\end{tabularx}
\end{table}

\subsection{Benchmarking}
In this section, we will show that DEM outperforms other classical estimators from control systems with the least MSPE for five quadrotor flight experiments in wind. The estimators under consideration are Prediction Error Minimization (PEM), Subspace method (SS) and Expectation Maximization (EM). The System Identification toolbox from MATLAB was used for SS (\textit{n4sid()}) and PEM (\textit{pem()}) methods, whereas an EM algorithm implementation for state space model was written based on \cite{cara2014using}. PEM was initialized using the solutions of SS. The data from experiment 2 was used to learn the state space model of the quadrotor for all methods. Fig. \ref{fig:y_pred_benchmark} shows the results of the n step ahead output prediction on the unseen test data (grey background) using the model learned by all methods. All predictions tend to follow the trend of the measured output (red curve). The prediction accuracy of different methods in Fig. \ref{fig:y_pred_benchmark} is visualized in Fig. \ref{fig:pred_error} using the n step ahead squared prediction error. It can be observed that DEM outperforms other methods with the least prediction error on unseen data. MSPE was used as the evaluation metric to compare the performance of DEM with other methods on all five flight data, and the results are shown in Fig. \ref{fig:MSPE_bench}. DEM outperforms other methods for all five experiments with minimum MSPE on unseen test data.




\begin{figure}[!b]
\centering
\includegraphics[scale=.38,trim={0.1cm 0 1cm 0},clip]{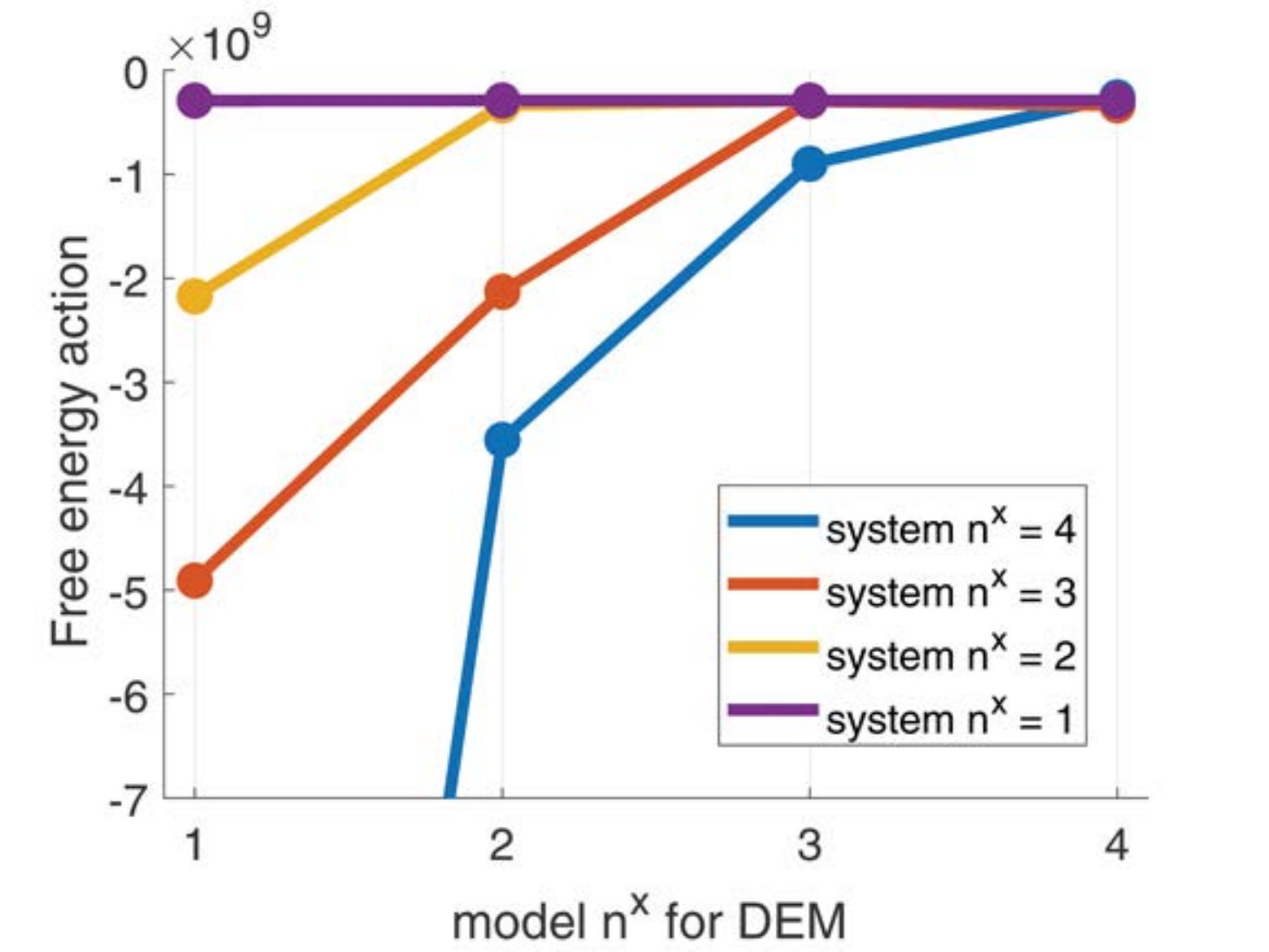}
\caption{The average $\Bar{F}$ of five experiments for different model orders. $\Bar{F}$ saturates when the robot brain's model order matches the real system order (4 for blue, 3 for red, 2 for yellow and 1 for violet). }
\label{fig:model_nx}
\end{figure}

\subsection{Extended DEM for black-box estimation}
The previous sections use the known inputs, outputs and model order for the output prediction, which differs from the biological brain's perception that do not have access to the real inputs and model order. Therefore, in this section we unleash the full capability of DEM with unknown inputs, and then extend it by proposing a free energy objective based scheme to evaluate the model order for black box estimation. Since $\Bar{F}$ is the sum of prediction errors for $\tilde{x}$, $\tilde{v}$, $\theta$ and $\lambda$ and their entropies, it is intuitive for the correct model order (or above) to maximize $\Bar{F}$. In this section, we test this hypothesis for quadrotor flights.

We consider the linearlized model of the quadcopter dynamics given in \cite{benders2020ar} to derive 4 different LTI systems as given in Table \ref{tab:systems}. All systems are observable and controllable, and use motor pwm signals as the input. $y$, $\dot{y}$ $ \phi$ and $ \dot{\phi}$ were selected as states since they are the most influenced by the wind in $y$ direction, thereby generating colored process noise in data. We use the same algorithm setup in Section \ref{sec:alg_settings} except for an additional constraint for unknown input ($\eta^v = 0$ with low precision $P^v = e^2$), to run DEM for all five experiments. The converged values of $\Bar{F}$ for all experiments were recorded by assuming a model order of 1,2,3 and 4 for all systems. The average $\Bar{F}$ of five experiments with different model orders for all four systems is shown in Fig. \ref{fig:model_nx}. $\Bar{F}$ saturates when the model order matches the system order, proving that $\Bar{F}$ is an indicator for model order selection. We use this idea to extend the original DEM algorithm for complete black box estimation as given in Algorithm \ref{algo:extended_DEM}. It generates an internal model via free energy maximization to estimate $\tilde{x}, \tilde{v}, \theta, \lambda$ and $n^x$ that best explains the data.

\begin{algorithm}[]
\SetAlgoLined
Initialize priors $\eta = \{\eta^v,P^v,\eta^\theta,P^\theta,\eta^\lambda,P^\lambda\}$\;
Initialize brain's model order $n^x =0 $ \;
\While(\Comment{model order}){$\Bar{F}_{b}$ not converged}{
 Initialize $a \xleftarrow{} 1$ and $\Bar{F}_a \xleftarrow{} -\infty$; \\ 
 $n^x \xleftarrow{} n^x + 1$; \Comment{increment model order} \\
 \While{$\theta$ not converged}{
    \For{t = 0:$\Delta t$:T}{ 
    $\tilde{x}(t), \tilde{v}(t) \xleftarrow{}$ D\_STEP($\tilde{y}(t),\theta,\lambda,n^x,\eta$)\; 
    }
    \While(){$\lambda$ not converged}{
        $\lambda  \xleftarrow{}$ M\_STEP($\tilde{y},\tilde{x}, \tilde{v},\theta, \lambda, n^x,\eta$) \;
    }
  $\Bar{F}_a \xleftarrow{} \text{Equation } \eqref{eqn:F_action_opt} $ \Comment{$\Bar{F}$ at optimal precision} 
  \If(\Comment{update $\theta$ if $\Bar{F}$ increased}){$\Bar{F}_{a}>\Bar{F}_{a-1}$  } 
   { 
    $\theta  \xleftarrow{}$ E\_STEP$(\tilde{y},\tilde{x}, \tilde{v},\theta, \lambda, n^x,\eta)$ \; 
   } 
    $a \xleftarrow{} a+1$ \;
 }
 $\Bar{F}_{b} = \Bar{F}_{a-1}$\; 
 }
 \caption{Extended DEM - black box estimation}
 \label{algo:extended_DEM}
\end{algorithm}

\section{CONCLUSIONS}
Robust algorithms for robot perception is still an open challenge, which could be solved by a brain-inspired algorithm. We take a step towards applying one such method (DEM) to real robots. We introduced a DEM based perception and system identification scheme for accurate output predictions during uncertainties (unmodelled dynamics), and validated it through real experiments on a quadcopter. We demonstrated its superior performance through its minimum MSPE for 150 step ahead output prediction, when compared to estimators like SS, PEM and EM. The usefulness of generalized coordinates in providing additional (derivative) information for estimation during unmodelled dynamics (wind) was demonstrated through the decrease in MSPE. Based on the results, the original DEM algorithm was extended for model order selection during complete black box estimation. The main disadvantage of DEM is it's higher computational complexity induced by generalized coordinates. Moreover, the noise smoothness needs to be known $\textit{apriori}$. The future research will address these issues.


\addtolength{\textheight}{-12cm}   


\section*{ACKNOWLEDGMENT}
We would like to thank Dennis Benders for his involvement in the data acquisition, as a part of his masters thesis.

\bibliographystyle{IEEEtran}
\footnotesize
\bibliography{references/main}

\end{document}